\documentclass{article}
\usepackage{spconf,amsmath,graphicx}
\usepackage{xspace}
\usepackage{multirow}
\usepackage{amssymb}
\usepackage{caption,subcaption}
\usepackage{url}
\usepackage{booktabs}

\usepackage{color}
\newcommand{\fnote}[1]{{\color{blue} \bf #1 \color{blue}}}

\newcommand{\knoteJP}[1]{{\color{blue} \bf #1 \color{black}}}
\newcommand{\bnote}[1]{{\color{red} #1 \color{black}}}
\newcommand{\mnote}[1]{}

\newcommand{\snote}[1]{}
\newcommand{\kcut}[1]{}
\newcommand{\ocut}[1]{}
\newcommand{\scut}[1]{}
\newcommand{\jptext}[1]{{\color{blue} \bf #1 \color{black}}}

\newcommand{\figcaption}[1]{\def\@captype{figure}\caption{#1}}
\newcommand{\tblcaption}[1]{\def\@captype{table}\caption{#1}}

\ifx\pdfoutput\undefined
\else
 \usepackage{CJKutf8}
 \renewcommand{\fnote}[1]{}
 \renewcommand{\knoteJP}[1]{}
 \renewcommand{\mnote}[1]{}
 \renewcommand{\bnote}[1]{}
 \renewcommand{\jptext}[1]{}
\fi

\newcommand{\argmin}{\mathop{\rm argmin}\limits}

\newcommand{\figref}[1]{{Fig.~\ref{fig:#1}}}
\newcommand{\tabref}[1]{{Table \ref{tab:#1}}}

\newcommand{\secref}[1]{Sec.~\ref{sec:#1}}

\makeatletter
\DeclareRobustCommand\onedot{\futurelet\@let@token\@onedot}
\def\@onedot{\ifx\@let@token.\else.\null\fi\xspace}

\def\ie{\emph{i.e}\onedot}

\def\etal{\emph{et al}\onedot}
\makeatother

\title{PoseRN: A 2D pose refinement network for bias-free multi-view \\3D human pose estimation}
\name{Akihiko Sayo$^{\star}$ \quad Diego Thomas$^{\star}$ \quad Hiroshi Kawasaki$^{\star}$ \quad Yuta Nakashima$^{\dagger}$ \quad Katsushi Ikeuchi$^{\ddagger}$}

\address{$^{\star}$ Kyushu University, Japan, 
  ~~~$^{\dagger}$ Osaka University, Japan, 
  ~~~$^{\ddagger}$ Microsoft Corp, USA}

\begin{document}

\maketitle

\begin{abstract}

We propose a new 2D pose refinement network that learns to predict the human bias in the estimated 2D pose. There are biases in 2D pose estimations that are due to differences between annotations of 2D joint locations based on annotators' perception and those defined by motion capture (MoCap) systems. These biases are crafted into publicly available 2D pose datasets and cannot be removed with existing error reduction approaches.
Our proposed pose refinement network allows us to efficiently remove the human bias in the estimated 2D poses and achieve highly accurate multi-view 3D human pose estimation.
\end{abstract}
\vspace{-0.2cm}
\section{Introduction}\label{sec:intro}
\vspace{-0.2cm}

Early works on 3D human pose estimation have using convolutional neural networks (CNN)~\cite{kolotouros2019learning, cheng2019occlusion, Sun2018integral, Martinez_2017_ICCV,zhou2017towards, NibaliHMP19} have focused on using a single image as input.
This is an ill-posed problem and as a consequence, such CNN-based methods rely much on the knowledge from the dataset. However, unlike 2D human pose datasets, publicly available MoCap-based 3D human pose datasets only contain images of a limited number of persons taken in a controlled environment and with tight clothes.

To take advantage of the publicly available large 2D datasets, methods were proposed to estimate 3D human poses from multi-view 2D images~\cite{iskakov2019learnable, multiviewpose, Joo2017panoptic, zhou2019hemlets, pavlakos2017harvesting, tome2018rethinking}. 
The core idea is to first estimate 2D poses in each view by using a 2D pose estimation network and then triangulate them to generate a 3D pose. To cope with noise in the 2D pose estimate two strategies exist: (1) defining a robust 3D triangulation algorithm~\cite{iskakov2019learnable, Joo2017panoptic, zhou2019hemlets} or (2) refining the 2D pose estimates prior to doing the 3D triangulation~\cite{multiviewpose, pavlakos2017harvesting, tome2018rethinking}.

\begin{figure}[t]
  \centering
  \includegraphics[width=0.8\linewidth]{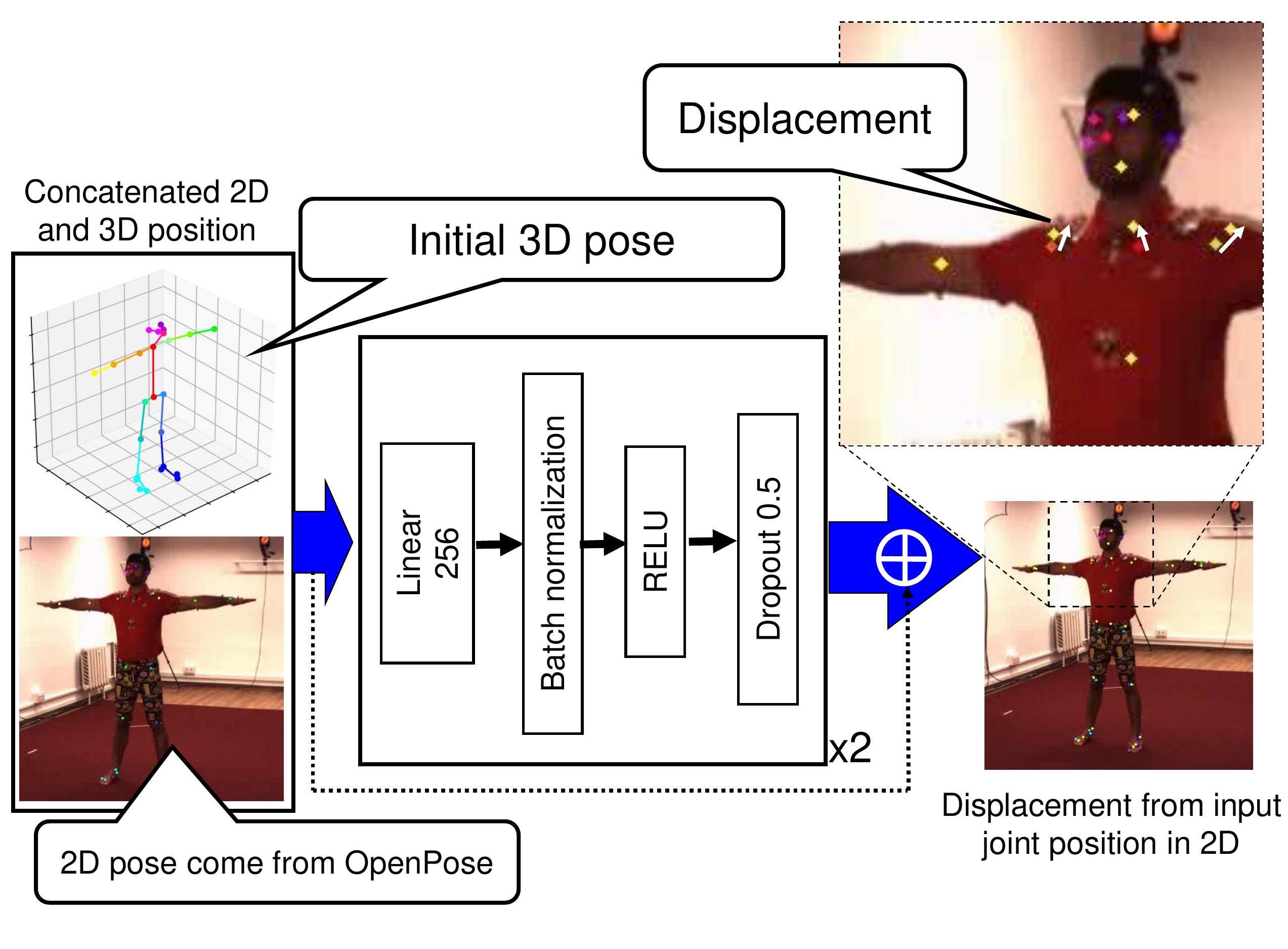}
  \vspace{-0.3cm}
  \caption{Network structure of pose refinement network (PoseRN).}
  \label{fig:PoseRN}
  \vspace{-0.3cm}
\end{figure}

In this work, we take the second approach. We observed in the commonly used datasets like Human3.6M \cite{ionescu2013human3} that there are consistent errors between projections of 3D poses and the corresponding 2D estimates (or even annotations).
We reason that errors in 2D pose estimates do not come solely from computational errors but also come from consistent displacements due to human perception bias that depends on the viewpoint and the pose. Unlike computational errors, biases cannot be removed by using averaging strategies. 

We propose a new 2D pose refinement network that leverages both 2D and 3D information to un-bias the initial 2D pose estimates and achieve accurate 3D pose estimation. Our core idea is to build a network, dubbed \textit{PoseRN}~(\ref{fig:PoseRN}), that learns to predict the 2D bias between the initial multi-view 2D poses and the anatomical ones defined by the (few) available MoCap-based 3D datasets. The 2D human biases depend on the camera viewpoint and the human pose. We reason that the camera viewpoint information is contained into the estimated 2D pose, while the human pose information is contained in the initial 3D pose estimate. 

In summary, our main contributions are (1) a marker-less multi-view 3D pose estimation method, which outperforms previous methods; (2) a pose-refining network, PoseRN, to rectify human perceptional 2D annotations to anatomical ones, which is defined by MoCap systems.

\begin{figure*}[t]
\centering
\begin{tabular}{ccccccc}
    &  S9       &  S11           & S1             & S6              & S5            \\
    & ``Posing''  & ``Directions 1'' & ``Discussion 1'' & ``Phoning 1''     & ``Discussion 3''\\
    &  Pelvis   &  Right Hip     & Left Hip       & Right Shoulder  &  Left shoulder\\
    \rotatebox[origin=l]{90}{\hspace{0.2cm}OpenPose} &
    \includegraphics[width=0.13\linewidth]{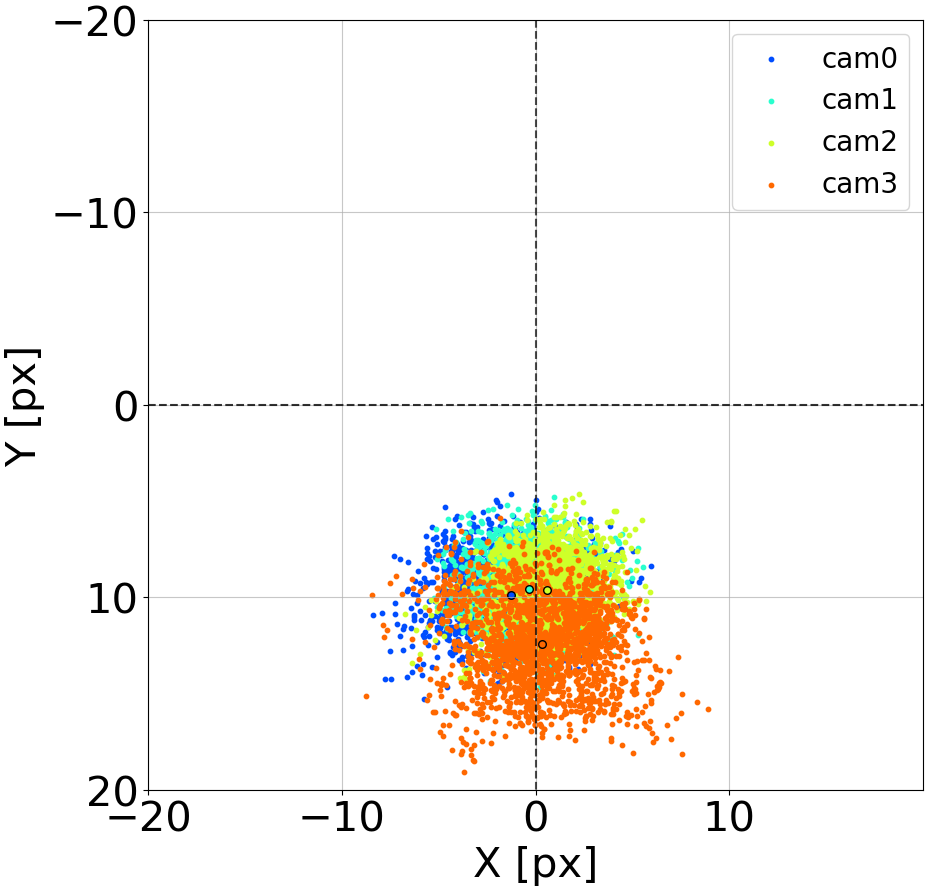} &
    \includegraphics[width=0.13\linewidth]{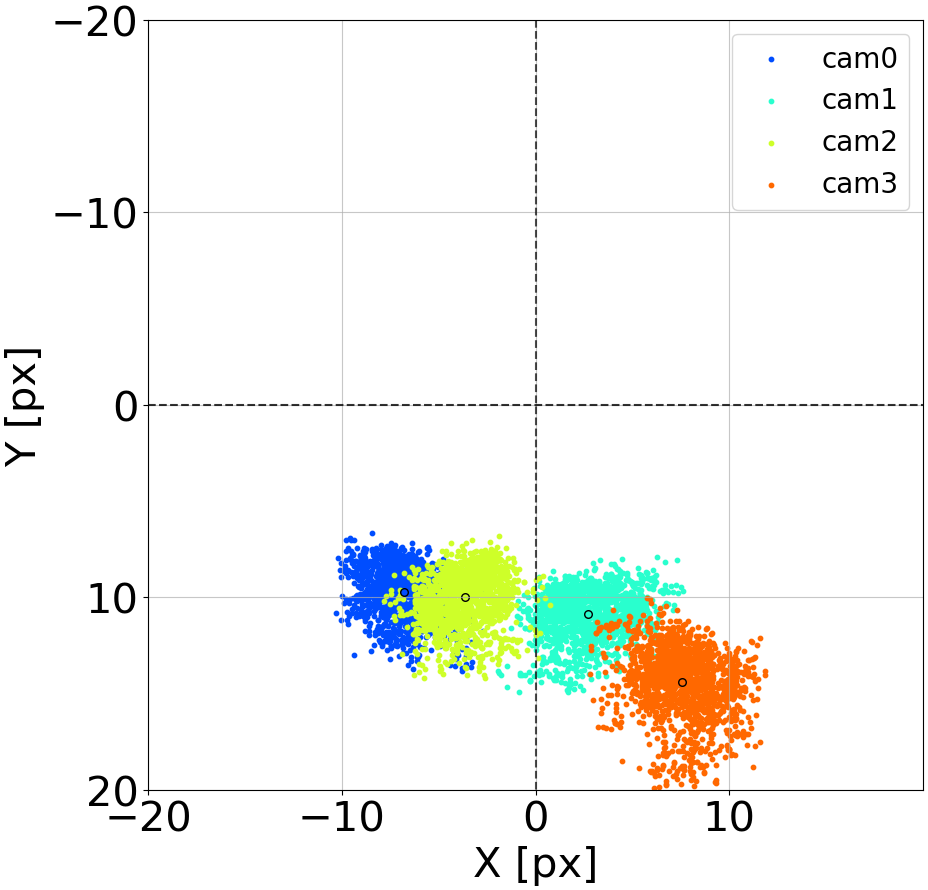} &
    \includegraphics[width=0.13\linewidth]{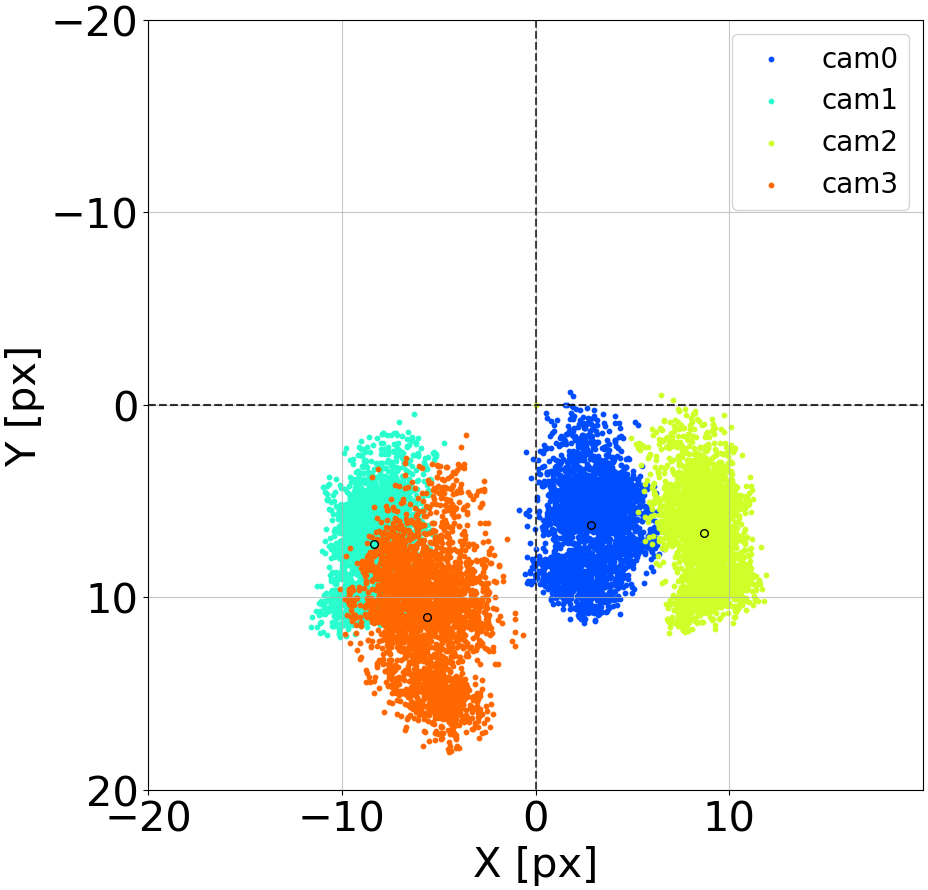}&
    \includegraphics[width=0.13\linewidth]{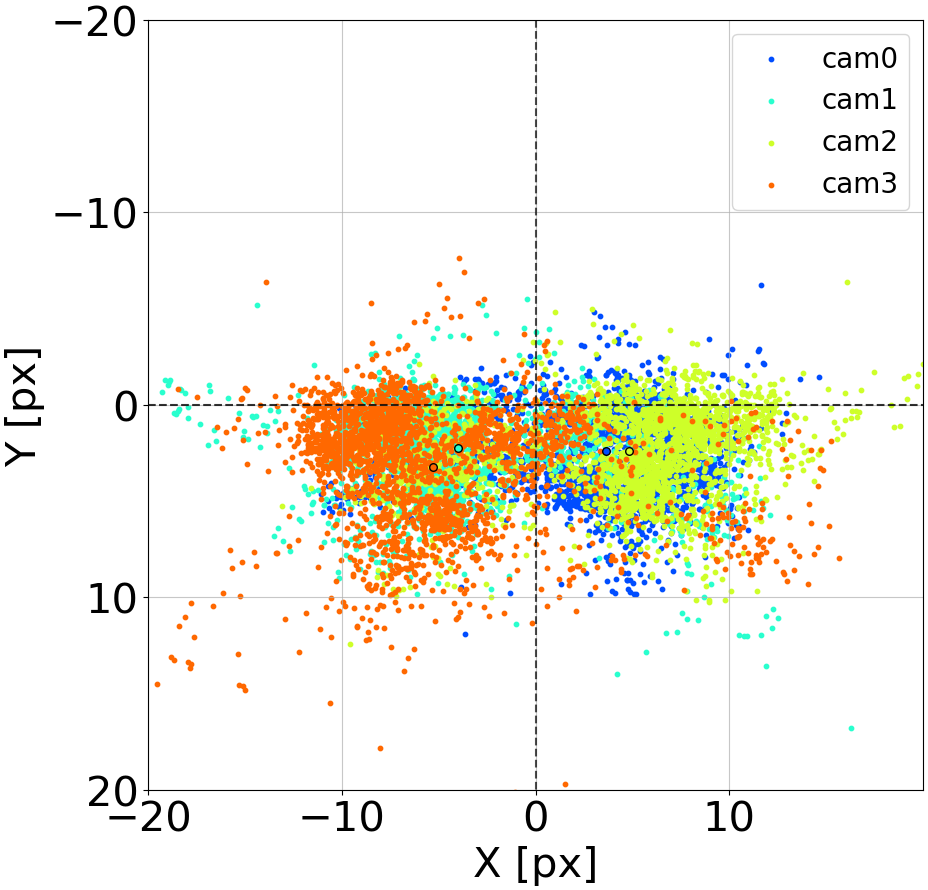}&
    \includegraphics[width=0.13\linewidth]{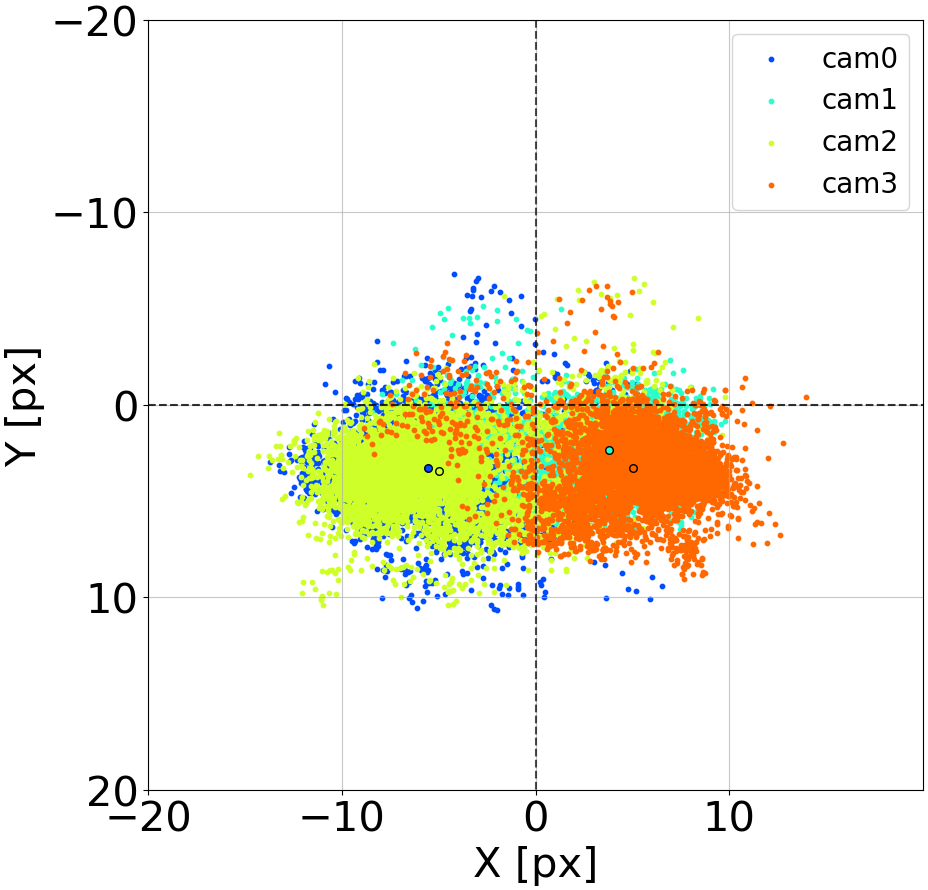}\\
    \rotatebox[origin=l]{90}{\hspace{0.2cm}PoseRN} &
    \includegraphics[width=0.13\linewidth]{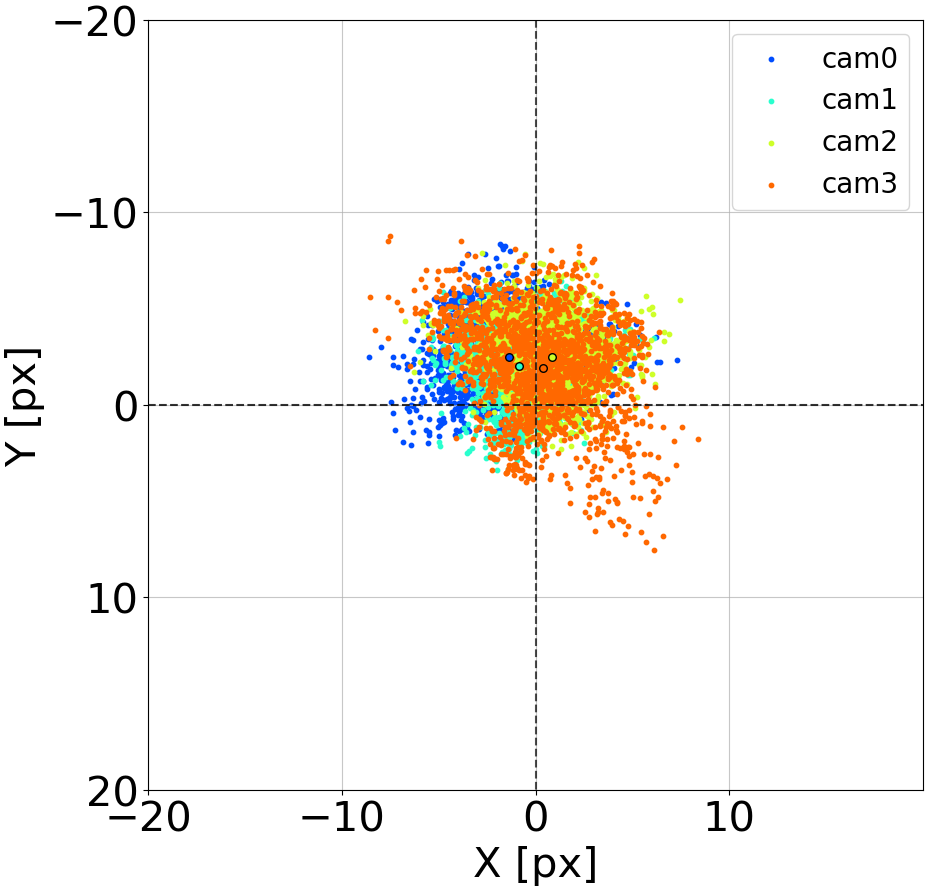} &
    \includegraphics[width=0.13\linewidth]{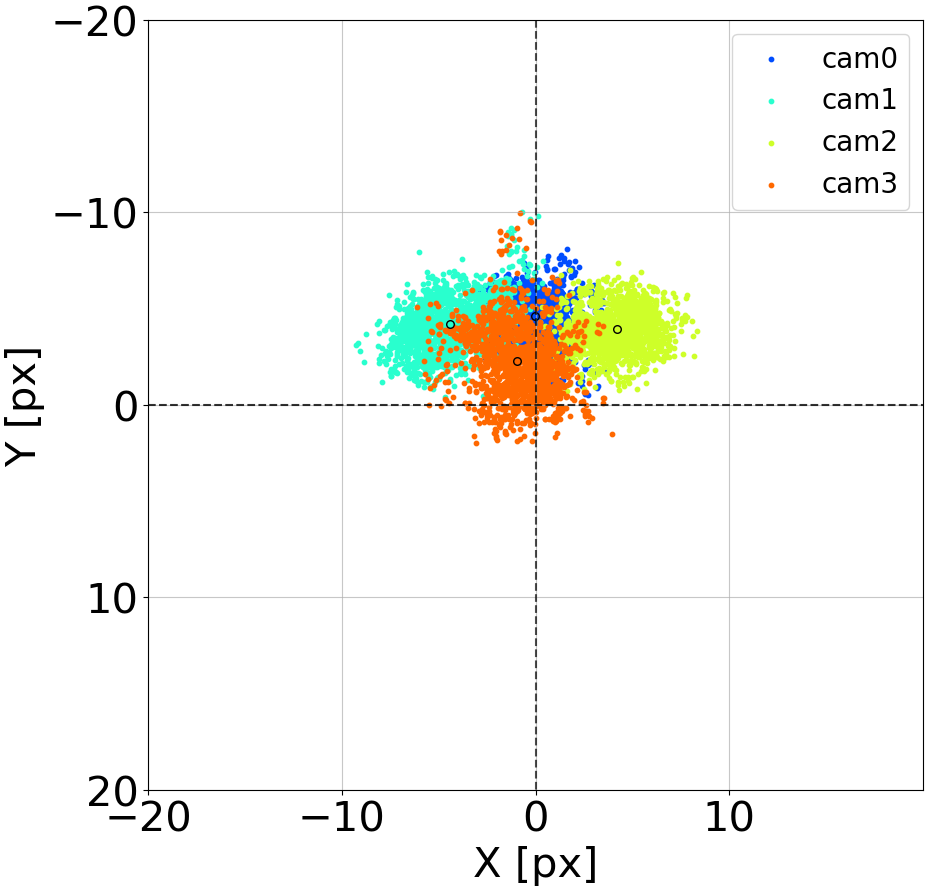} &
    \includegraphics[width=0.13\linewidth]{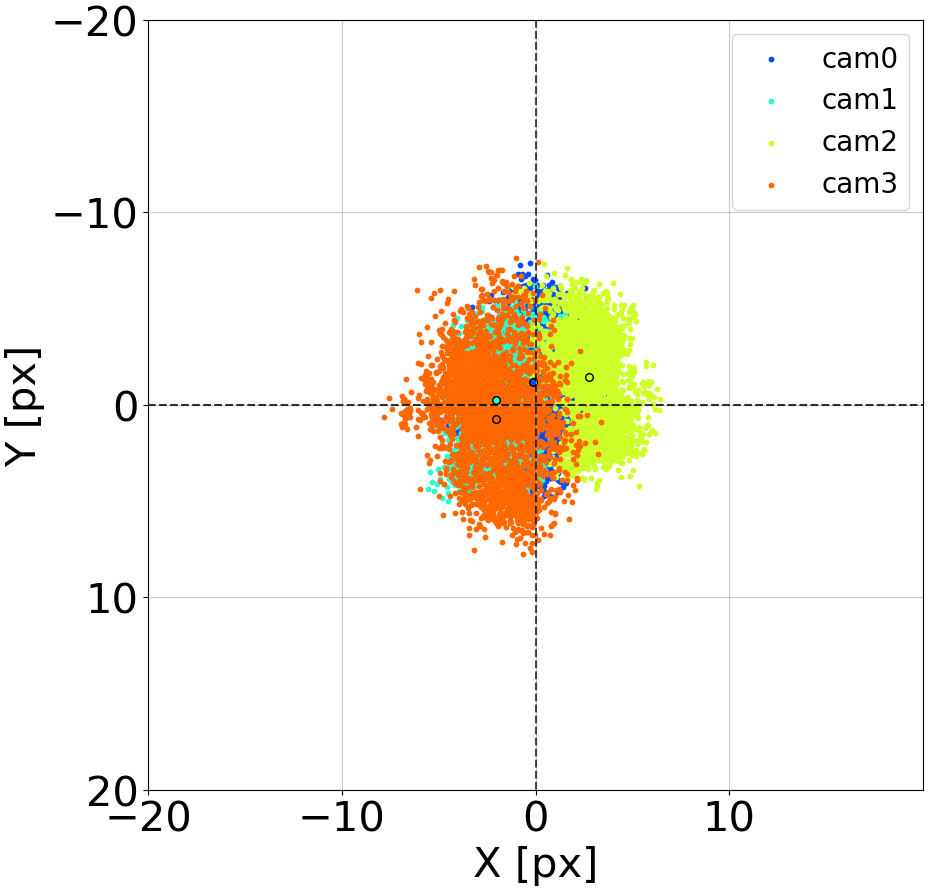}&
    \includegraphics[width=0.13\linewidth]{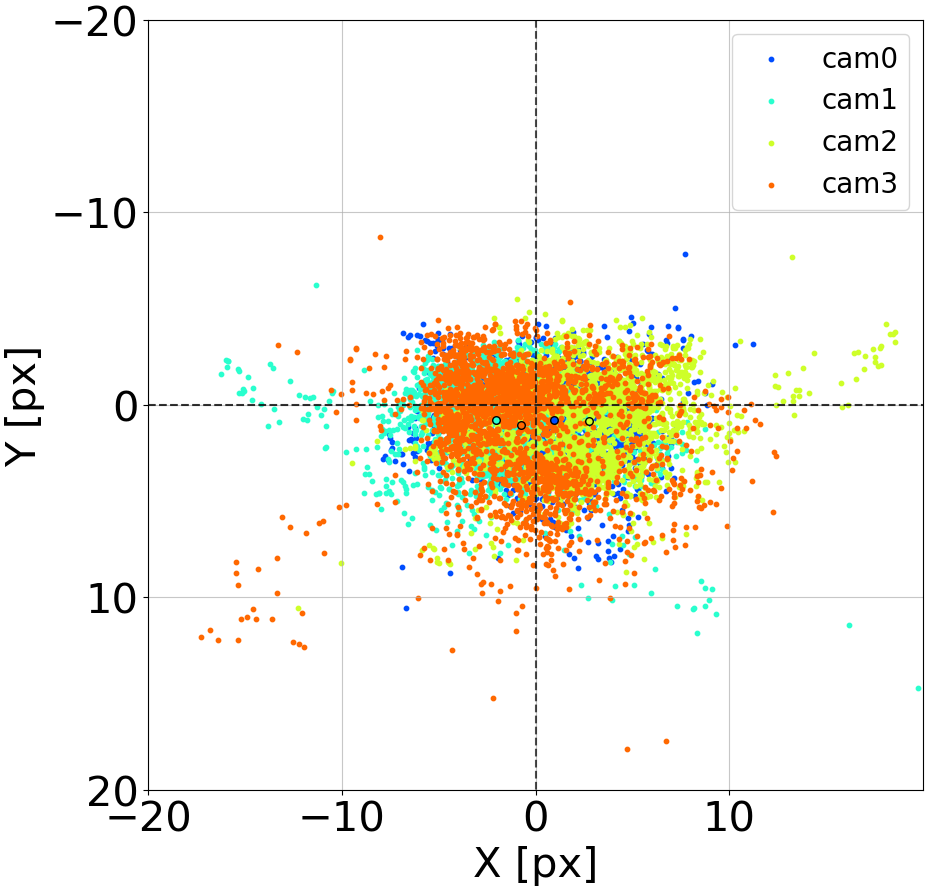}&
    \includegraphics[width=0.13\linewidth]{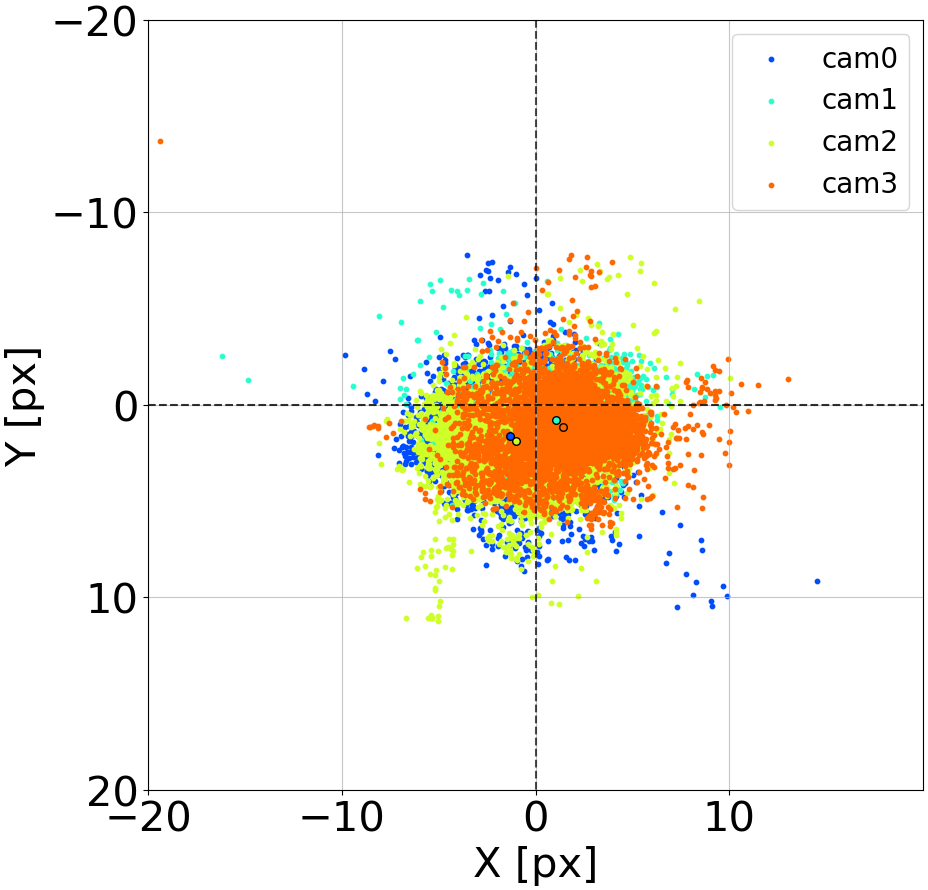}\\
\end{tabular}
\vspace{-0.3cm}
\caption{Analysis of the bias between 3D pose annotations Human3.6M and 2D pose estimates by OpenPose (top) and PoseRN (bottom). Each scatter plot shows the displacements from GT 2D joint position. Joint positions in the same color represents those in the same camera. }
\label{fig:analysis_bias}
\vspace{-0.3cm}
\end{figure*}

\vspace{-0.2cm}
\section{Biases in 2D pose annotations}
\label{sec::bias}
\vspace{-0.2cm}

The 3D poses obtained by triangulation based on estimated 2D poses in multi-view images, in general, are not accurate. 
One of the major factors that introduce such errors is that humans cannot make pixel-accurate annotations. This type of error behaves more like variance in annotations and can be alleviated by averaging over multiple annotations.
Aside from this variance, there are inevitable errors between 2D human pose annotations that come from the difference of joint definitions for 3D and 2D joint annotations.

The 3D poses in the datasets, such as Human3.6M \cite{ionescu2013human3}, used to train and evaluate 3D human pose estimation methods are obtained with a MoCap system. As a consequence, each dataset has its own marker-based definition of joints. Such joint positions may be computed by averaging multiple marker positions associated with a certain joint; therefore, the obtained joint positions can be inside the human body. 

Meanwhile, the dataset used for training 2D pose estimation employed human annotators. When manually annotating a 2D image, an annotator is asked to click on the pixels that should be the 2D re-projection of the 3D joint. However, the 3D joints are inside the body, so the annotator can only guess the 2D re-projection location by looking at the surface of the body.
This means that a 3D joint position is less likely to be on the ray formed by the camera center and the corresponding annotated 2D joint position.  

We reason that humans have a similar perception of the human body and that the errors made when guessing 2D joint locations tend to be similar for different persons. As a consequence, there is a bias in 2D pose annotation datasets that depends on both camera pose and human body pose. This can be seen in the top row of  Fig.~\ref{fig:analysis_bias}.

\begin{figure}[t]
    \centering
    \includegraphics[width=0.95\linewidth]{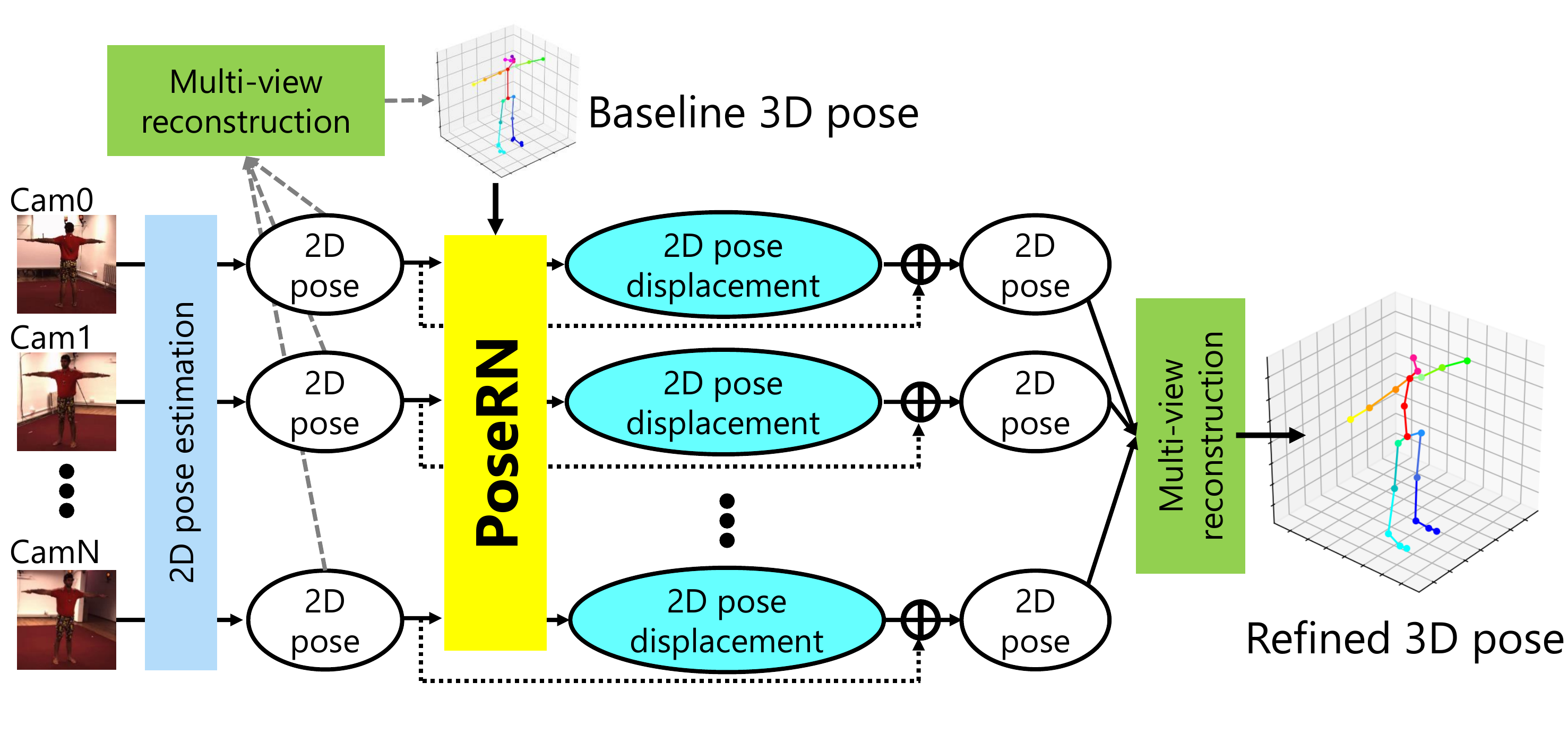}
    \vspace{-0.5cm}
    \caption{Overview of our proposed two-pass 3D pose estimation method. }
    \label{fig:overview}
    \vspace{-0.3cm}
\end{figure}

\vspace{-0.2cm}
\section{3D pose estimation with 2D pose de-biasing} \label{sec:3Dpose}
\vspace{-0.2cm}
\subsection{De-biasing 2D poses}
\label{sec:debias}
\vspace{-0.1cm}

As discussed in the previous section, the displacements in 2D pose estimates may be described by 1) the discrepancy in the definitions of joints for 3D and 2D pose annotations, 2) human poses, and 3) camera positions. More specifically, letting $\hat{J}^\text{3D}$ and $\hat{J}^\text{2D}$ denote a 3D joint position annotation and its corresponding 2D joint position annotation in a certain camera image, we assume that the relationship between $\hat{J}^\text{3D}$ and $\hat{J}^\text{2D}$ can be described by:
\begin{equation}
    \Pi(\hat{J}^\text{3D}|M) = \hat{J}^\text{2D} + \beta(\hat{J}^\text{3D}, M) + \epsilon. \label{eq:bias_2}
\end{equation}
where $M$ is the camera parameter for the image, $\Pi(\cdot|M)$ the projection and $\epsilon$ is some noise.
$\beta(\hat{J}^\text{3D}, M)$ is the bias between the projection of $\hat{J}^\text{3D}$ and $\hat{J}^\text{2D}$.

Our PoseRN ideally learns this $\beta$ to de-bias the estimated 2D poses. 
We use 3D pose estimates $J^\text{3D} = \{J_j^\text{3D}|j \in \mathcal{J}\}$ and 2D pose estimates $J^\text{2D} = \{J_{cj}^\text{2D}|c \in \mathcal{C}, j \in \mathcal{J}\}$ as input to PoseRN, where $J_j^\text{3D}$ is the 3D position of joint $j$ of the 3D pose estimate; $J_{cj}^\text{2D}$ the 2D position of joint $j$ in camera $c$ of the 2D pose estimates; $\mathcal{J}$ and $\mathcal{C}$ are the index sets for joints and cameras.

Similarly to \cite{Martinez_2017_ICCV}, our network consists of two linear layers followed by batch normalization, ReLU, and dropout~(\figref{PoseRN}).
Before training, all 3D pose annotations and 2D poses estimated from the input images are translated so that the pelvis joint coincides with the origin of the local coordinate system. 

We employ the MSE loss and minimize the error between real biases and predicted biases, given by:
\begin{equation}
    \ell = \mathbb{E}[\|\beta_{cj} - \text{PoseRN}(J^\text{3D}, J^\text{2D}_{c})\|^2],
\end{equation}
where $\beta_{cj} = J^\text{2D}_{cj} - \Pi(\hat{J}^\text{3D}_j|M_c)$ based on Eq.~(\ref{eq:bias_2}), $\text{PoseRN}$ is the output of the network, $J^\text{2D}_{c}$ is the set of 2D joint positions in camera $c$.
Both 3D joint positions and 2D joint positions are concatenated into respective vectors to form the input to the network. At test time, we de-bias the 2D pose by subtracting the predicted bias from 2D pose estimate $J^\text{2D}_{c}$. 

\vspace{-0.2cm}
\subsection{multi-view 3D pose reconstruction}
\label{sec:MV3Dpose}

Figure \ref{fig:overview} shows the overview of our method. Specifically, the pipeline consists of 1) 2D poses estimation from multi-view images, 2) 
initial 3D pose estimation by multi-view triangulation,
3) 2D poses de-biasing by PoseRN, and 4) final 3D pose estimation from de-biased 2D poses.

\textbf{2D pose estimation}:
We firstly estimate 2D human pose $J^\text{2D}$ in multi-view images independently. We use OpenPose \cite{cao2018openpose} as our 2D pose estimator, in which each joint position has the probability of the joint being at the pixel position. This probability, $w_{cj}$, corresponding to $J_{cj}^{2D}$ can be in turn viewed as the confidence of the joint position.  

\textbf{Initial 3D pose estimation}:
On multi-view system, the cameras are calibrated and thus the camera parameter $M_c$ of camera $c$ is known.  We employ multi-view geometry to compute 3D pose $J^\text{3D}$ from multiple 2D pose estimates $J^\text{2D}$. Specifically, the initial 3D pose is obtained by 
\begin{equation}
J^\text{3D} = \argmin_{\bar{J}^\text{3D}} \sum_{c \in \mathcal{C}} \sum_{j \in \mathcal{J}} \rho(w_{cj}\|\Pi(\bar{J}^\text{3D}_{j}|M_c) - J^\text{2D}_{cj}\|),
\label{equ:energy}
\end{equation}
where $\rho(\cdot)$ is the Huber loss. Note that we use confidence $w_{cj}$ as weight because a joint with a low confidence may imply a significant error in the 2D pose estimation. When the 2D positions of joint $j$ are not estimated in images of all camera or are estimated only in a single image, we exclude joint index $j$ from $\mathcal{J}$ to ignore the joint. 

\textbf{2D pose de-biasing by PoseRN}:
PoseRN takes the initial 3D pose estimate $J^\text{3D}$ and 2D pose estimates $J^\text{2D}$ as input and predicts the bias between 3D pose and 2D poses. As mentioned in Section \ref{sec:debias}, we de-bias the 2D pose estimates by
\begin{equation}
    \tilde{J}^\text{2D}_{cj} = J^\text{2D}_{cj} + \text{PoseRN}_{cj}(J^\text{3D},J^\text{2D}),
\end{equation}
where $\text{PoseRN}_{cj}$ is predicted bias for joint $j$ in camera $c$. 

\textbf{Final 3D pose estimation}:
In order to compute the final 3D pose estimate $\tilde{J}^\text{3D}$, we use Eq.~(\ref{equ:energy}) but the de-biased 2D estimates $\tilde{J}^\text{2D}$ are used instead of $J^\text{2D}$. The output of our method is thus $\tilde{J}^\text{3D}$.

For PoseRN, we preprocessed the 2D poses from OpenPose~\cite{cao2018openpose}. They are re-scaled by the smaller length of the width or height of the input image and a third of the length of the spine. This scaling parameter is determined empirically so that the distributions of the joints coordinates in the estimated 2D poses and 3D poses become similar. We trained PoseRN using the Adam optimizer
with a learning rate of $0.001$ for $20$ epochs. 

\vspace{-0.2cm}
\section{Experiments}
\vspace{-0.2cm}
\begin{table*}[t]
    \centering
    \caption{MPJPE in $mm$ on MPI-INF-3DHP~\cite{mono-3dhp2017} (left) and TotalCapture~\cite{Trumble:BMVC:2017} (right).}
    \vspace{-0.3cm}
    \begin{tabular}{ccccccccc}
        \toprule
        & S1 & S2 & \multicolumn{3}{c}{S1} & \multicolumn{3}{c}{S4} \\
        \cmidrule(lr){2-2}\cmidrule(lr){3-3}\cmidrule(lr){4-6}\cmidrule(lr){7-9}
        & Seq1 & Seq2 & A3 & FS3 & W2 & A3 & FS3 & W2 \\
        \midrule
        Iskakov \etal~\cite{iskakov2019learnable} & 129.47 & 101.39 & 116.43 & 97.87 & 110.45 & 83.01 & 69.06 & 67.45\\
        Multi-view reconstruction & 64.09 & 53.13 & 50.02 & 59.92 & 58.69 & 50.52 & 63.41 & 45.07\\
        \textbf{Ours(PoseRN)} & 51.97 & 58.97 & 45.84 & 56.76 & 53.29 & 49.62 & 65.53 & 45.81\\
        \bottomrule
    \end{tabular}
    \label{tab:mpi-inf-3dhp}
\vspace{-.2cm}
\end{table*}

\begin{figure}[t]
    \centering
    \begin{tabular}{cccccc}
    \vspace{-0.1cm}
    \includegraphics[width=0.17\linewidth]{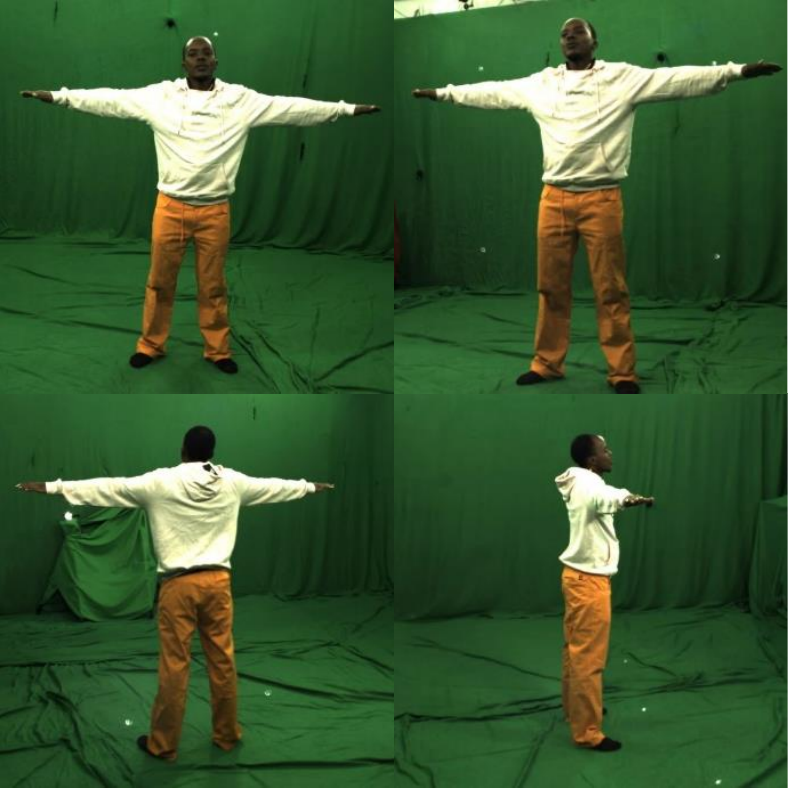}&
    \includegraphics[width=0.17\linewidth]{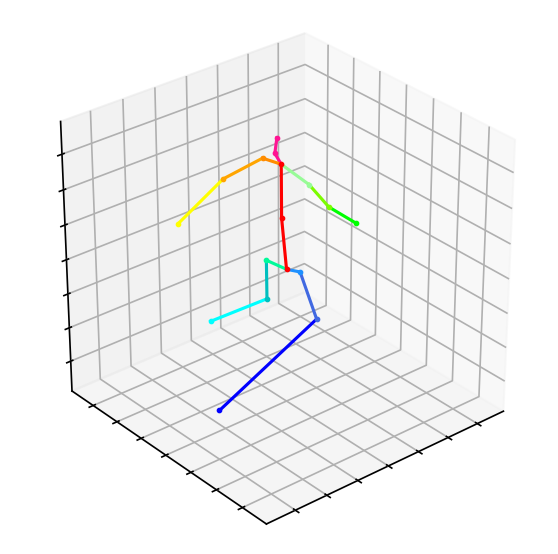}&
    \includegraphics[width=0.17\linewidth]{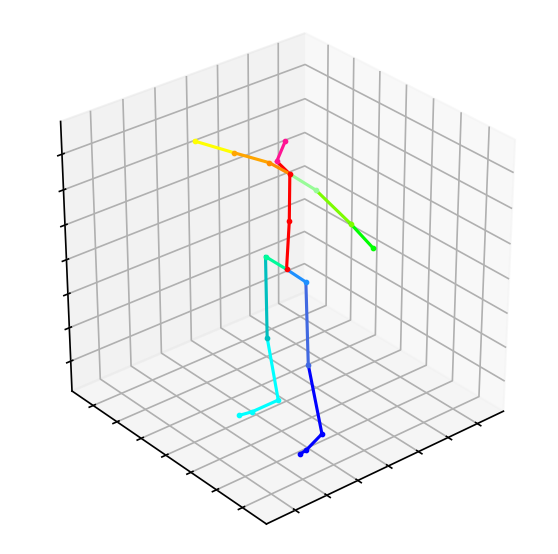}&
    \includegraphics[width=0.17\linewidth]{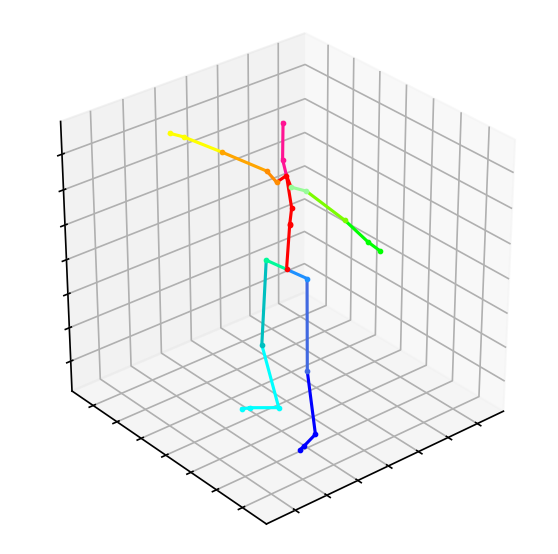} \\

    \includegraphics[width=0.17\linewidth]{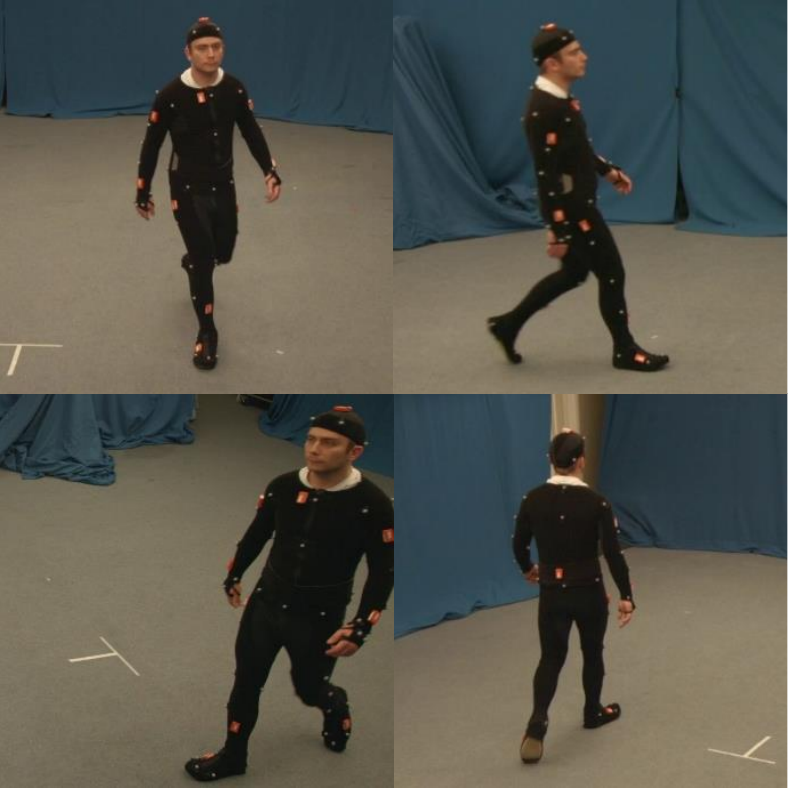} &
    \includegraphics[width=0.17\linewidth]{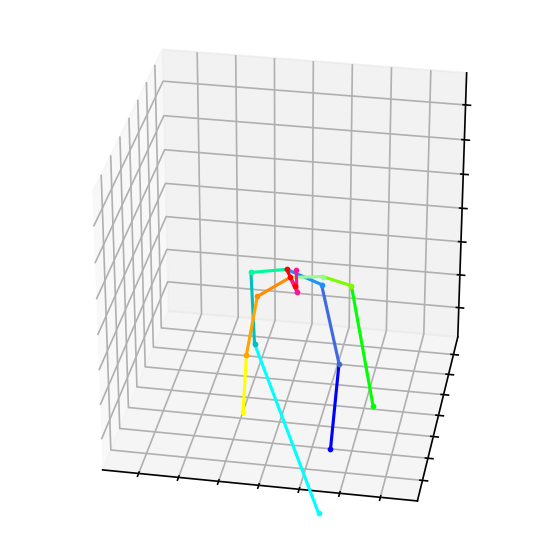} &
    \includegraphics[width=0.17\linewidth]{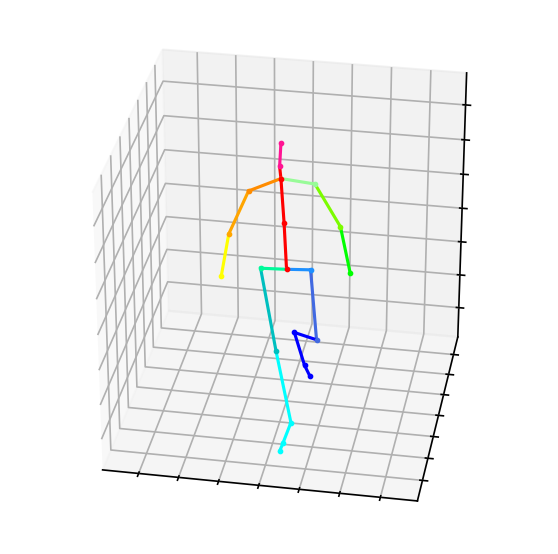} &
    \includegraphics[width=0.17\linewidth]{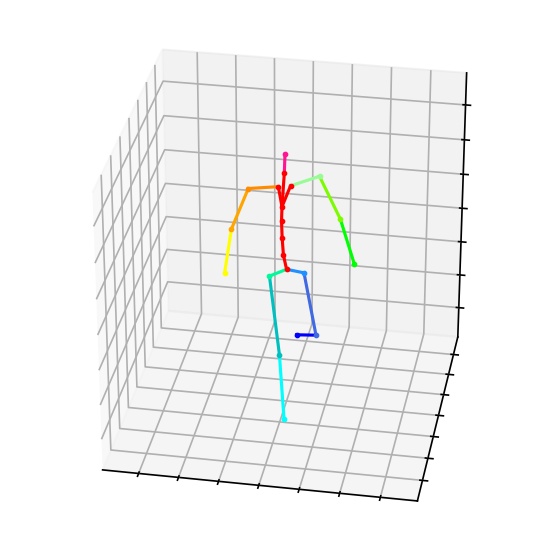} \\
    
    \end{tabular}
    \vspace{-0.2cm}
    
    \caption{Qualitative results on MPI-INF-3DHP \cite{mono-3dhp2017} and TotalCapture \cite{Trumble:BMVC:2017}. From left to right: input images, Iskakov \etal \cite{iskakov2019learnable}, ours, and ground-truth 3D pose annotations. 
    }
    \vspace{-0.1cm}
    \label{fig:qual_cross_dataset}
    \vspace{-0.3cm}
\end{figure}

We evaluate the performance of our proposed method with three experiments. 
First, we evaluate the ability of PoseRN to reduce the bias using Human3.6M dataset.
The second experiment evaluates the generalization performance in comparison with state-of-the-art Learnable Triangulation \cite{iskakov2019learnable}. We trained both methods on the Human3.6M dataset \cite{ionescu2013human3} and tested them on other 3D human pose datasets, \ie,  MPI-INF-3DHP~\cite{mono-3dhp2017} and TotalCapture~\cite{Trumble:BMVC:2017}. 
In the third experiment, we compare our proposed multi-view 3D pose estimation method with other multi-view 3D human pose estimation methods quantitatively.
We employ the mean-per-joint position error (MPJPE) as our evaluation criterion, which is the average Euclidean distance between an estimated joint position to the ground-truth joint position in millimeters.

Note that for all the methods evaluated in our experiments, we discarded the input data for which OpenPose totally fails.
We applied this elimination to all methods. Also, we re-trained all previous methods with the same datasets used for our proposed method for a fair comparison.  

\vspace{-0.2cm}
\subsection{Qualitative evaluation of PoseRN}
\vspace{-0.1cm}
Figure \ref{fig:analysis_bias} shows displacements of 2D pose estimates by OpenPose and de-biased 2D pose estimates by PoseRN in some example multi-view videos. More specifically, each dot in the figure represents $J^\text{2D}_{cj} - \Pi(J^\text{3D}_{j}|M_c)$ for OpenPose and $\tilde{J}^\text{2D}_{cj} - \Pi(J^\text{3D}_{j}|M_c)$ for PoseRN. We can see that some OpenPose's joint positions are clearly off from the origin, indicating the presence of the bias. In contrast, PoseRN successfully reduces this bias. The fact that the bias is predictable based on $J^\text{2D}_c$ and $J^\text{3D}$ may inductively support our assumption that the bias comes from the discrepancy in the joint definitions. 

\vspace{-0.2cm}
\subsection{Generalization to other dataset}\label{sec:exp_other_dataset}
\vspace{-0.1cm}
To evaluate the generalization performance of our method, we compare it with Iskakov \etal \cite{iskakov2019learnable} and vanilla multi-view triangulation in Eq.~(\ref{equ:energy}). Ours and Iskakov \etal's are trained on Human3.6M. All methods are tested on some excerpts from TotalCapture and the training split of MPI-INF-3DHP. Specifically, from MPI-INF-3DHP, we used Seq1 of S1 and Seq2 of S2.
From TotalCapture, we used $2$ subjects (s1 and s4) and $3$ actions for each subject (\textit{acting3}, \textit{freestyle3}, and \textit{walking2}). During the test on MPI-INF-3DHP, we measured the MPJPE on $12$ joints: Neck, L/R shoulder, L/R elbow, L/R wrist, Pelvis, L/R hip, L/R knee. We used the same joints except for Pelvis, L/R hip, and L/R knee, during the test on TotalCapture.

Note that \cite{iskakov2019learnable} takes the 2D bounding box of the person as input. We created the bounding boxes from the ground-truth foreground masks provided in both datasets and use them to crop the input images before estimating the 2D pose. Our proposed method has the advantage that it works directly on the 2D pose and does not require image input. As a consequence, we do not need the bounding box as input and our proposed method is insensitive to changes in the appearance of the person and its background. 

The quantitative results are shown in \tabref{mpi-inf-3dhp}. Our method outperformed \cite{iskakov2019learnable} for all the sequences. We carefully checked each frame and found that \cite{iskakov2019learnable} has severe errors when the subjects are wearing loose clothes with rich textures or when the sequence was captured by cameras that are placed far apart from the subject compared to Human 3.6M as shown in \figref{qual_cross_dataset}. These results show the generalizability of our method compared to SOTA (\ie, \cite{iskakov2019learnable}) in terms of changes in the appearance and viewpoint. 
In some sequences, the baseline multi-view method slightly outperformed our proposed method because some poses in these sequences were rarely seen in the training data.

In addition, PoseRN has the advantage that it successfully removed the annotation bias even when the subject wears totally different clothes and when the viewing directions are totally different from those of the training dataset.

\vspace{-0.1cm}
\subsection{Evaluation on Human3.6M}\label{sec:exp_3Dpose}
\vspace{-0.1cm}
\begin{table}[t]
    \centering
    \caption{Quantitative evaluation of the methods that do not use ground-truth 3D pose annotations in the training. All values except ours are adopted from the original papers. The number of the joints following each method name is used to measure MPJPE in each method.}
    \vspace{-0.2cm}
    \label{tab:exp_3Dpose_MV}
    \begin{tabular}{cccccccccccccccccc}
    \toprule
      Method  & Avg.~MPJPE (mm)\\ 
      \midrule
    Qiu~\etal~\cite{multiviewpose} (17 joints) & 43.0 \\
    Iskakov~\etal~\cite{iskakov2019learnable} (algebraic) (6 joints) & 36.0 \\
    Iskakov~\etal~\cite{iskakov2019learnable} (volumetric) (6 joints) & 34.0 \\
    Ours(MV optimization) (17 joints) & 40.0 \\
    \textbf{Ours(PoseRN)} (17 joints) & 38.4 \\
    \textbf{Ours(PoseRN)} (4 joints) & 29.6 \\
    \bottomrule
    \end{tabular}
    \vspace{-.5cm}
\end{table}

We quantitatively evaluated our proposed method with two recent methods \cite{iskakov2019learnable} and \cite{multiviewpose} on Human3.6M. \tabref{exp_3Dpose_MV} shows the results when all methods are tested on Human3.6M and trained on other datasets. We trained our proposed network on the MPI-INF-3DHP dataset \cite{mono-3dhp2017} (Seq1 and Seq2 of S1 and S2). 
MV optimization means that we just use the 2D pose estimator pre-trained on the 2D human pose dataset and conduct multi-view optimization as described in~\secref{MV3Dpose}. 

Here we assume that the joints positioned inside the body may have more bias than that of near the surface (wrist, ankle, etc...). To test this assumption, we have measured MPJPE for all estimated joints and subset of them. When we use all joints to measure, the MPJPE is 38.4mm and when use a subset joints~(neck, L/R elbow, spine), that is 29.6mm. 
These results confirmed our hypothesis and demonstrated the ability of PoseRN to reduce not only the gaussian errors but also the errors that come from the human perceptual bias.

Qiu \etal \cite{multiviewpose} first estimates the 3D human pose with a network pre-trained on MPII Human Pose Dataset \cite{andriluka14cvpr} and then fine-tune the pre-trained network on the estimated 3D poses. Iskakov~\etal~\cite{iskakov2019learnable} train their network on CMU Panoptic dataset~\cite{Joo2017panoptic}, using $27$ of the $31$ HD cameras. They test their network on Human3.6M and measure MPJPE for the subset of all joints (L/R elbows, L/R wrists, and L/R knees). Although the Panoptic dataset has more variety in terms of camera viewpoints and subjects' appearance than the MPI-INF-3DHP, the results of our method are more accurate than that of~\cite{iskakov2019learnable}. This confirms that our method (or PoseRN) requires only a few camera viewpoints and small variation on subjects during training to achieve accurate 3D human pose estimation. In addition, unlike Qiu \etal's \cite{multiviewpose}, our method can be applied to any camera setup. 

\vspace{-0.2cm}
\section{Conclusion}\label{conclusion}
\vspace{-0.2cm}
We introduced a simple yet effective method for multi-view 3D pose estimation that does not require any marker. 
We proposed PoseRN that allows us to learn the bias due to the difference of the joint definitions for 3D pose annotations and 2D pose annotations. 
Our experimental results demonstrated that our method has a clear advantage when the situations of training data and test data are totally different. 

\vspace{-0.4cm}
\section*{\centering \large Acknowledgment}
\vspace{-0.3cm}
This work was supported by JSPS/KAKENHI 20H00611, 18K19824, 18H04119 in Japan.
\vspace{-0.4cm}

\bibliographystyle{IEEEbib}
\bibliography{icip2021}

\begin{thebibliography}{10}

\bibitem{kolotouros2019learning}
Nikos Kolotouros, Georgios Pavlakos, Michael~J Black, and Kostas Daniilidis,
\newblock ``Learning to reconstruct 3d human pose and shape via model-fitting
  in the loop,''
\newblock in {\em Proceedings of the IEEE International Conference on Computer
  Vision}, 2019, pp. 2252--2261.

\bibitem{cheng2019occlusion}
Yu~Cheng, Bo~Yang, Bo~Wang, Wending Yan, and Robby~T Tan,
\newblock ``Occlusion-aware networks for 3d human pose estimation in video,''
\newblock in {\em Proceedings of the IEEE International Conference on Computer
  Vision}, 2019, pp. 723--732.

\bibitem{Sun2018integral}
X~Sun, B~Xiao, F~Wei, S~Liang, and Y~Wei,
\newblock ``Integral human pose regression,''
\newblock in {\em The IEEE European Conference on Computer Vision (ECCV)},
  2018, pp. 529--545.

\bibitem{Martinez_2017_ICCV}
Julieta Martinez, Rayat Hossain, Javier Romero, and James~J. Little,
\newblock ``A simple yet effective baseline for 3d human pose estimation,''
\newblock in {\em The IEEE International Conference on Computer Vision (ICCV)},
  Oct 2017.

\bibitem{zhou2017towards}
Xingyi Zhou, Qixing Huang, Xiao Sun, Xiangyang Xue, and Yichen Wei,
\newblock ``Towards 3d human pose estimation in the wild: a weakly-supervised
  approach,''
\newblock in {\em Proceedings of the IEEE International Conference on Computer
  Vision}, 2017, pp. 398--407.

\bibitem{NibaliHMP19}
Aiden Nibali, Zhen He, Stuart Morgan, and Luke Prendergast,
\newblock ``3d human pose estimation with 2d marginal heatmaps,''
\newblock in {\em {IEEE} Winter Conference on Applications of Computer Vision,
  {WACV} 2019, Waikoloa Village, HI, USA, January 7-11, 2019}, 2019, pp.
  1477--1485.

\bibitem{iskakov2019learnable}
Karim Iskakov, Egor Burkov, Victor Lempitsky, and Yury Malkov,
\newblock ``Learnable triangulation of human pose,''
\newblock in {\em International Conference on Computer Vision (ICCV)}, 2019.

\bibitem{multiviewpose}
Haibo Qiu, Chunyu Wang, Jingdong Wang, Naiyan Wang, and Wenjun Zeng,
\newblock ``Cross view fusion for 3d human pose estimation,''
\newblock in {\em International Conference on Computer Vision (ICCV)}, 2019.

\bibitem{Joo2017panoptic}
H~Joo, T~Simon, X~Li, H~Liu, T~Lan, L~Gui, S~Banerjee, T.S Godisart, B~Nabbe,
  I~Matthews, T~Kanade, S~Nobuhara, and Y~Sheikh,
\newblock ``Panoptic studio: A massively multiview system for social
  interaction capture,''
\newblock in {\em IEEE Transactions on Pattern Analysis and Machine
  Intelligence}, 2017.

\bibitem{zhou2019hemlets}
Kun Zhou, Xiaoguang Han, Nianjuan Jiang, Kui Jia, and Jiangbo Lu,
\newblock ``Hemlets pose: Learning part-centric heatmap triplets for accurate
  3d human pose estimation,''
\newblock in {\em Proceedings of the IEEE International Conference on Computer
  Vision}, 2019, pp. 2344--2353.

\bibitem{pavlakos2017harvesting}
Georgios Pavlakos, Xiaowei Zhou, Konstantinos~G Derpanis, and Kostas
  Daniilidis,
\newblock ``Harvesting multiple views for marker-less 3d human pose
  annotations,''
\newblock in {\em Proceedings of the IEEE conference on computer vision and
  pattern recognition}, 2017, pp. 6988--6997.

\bibitem{tome2018rethinking}
Denis Tome, Matteo Toso, Lourdes Agapito, and Chris Russell,
\newblock ``Rethinking pose in 3d: Multi-stage refinement and recovery for
  markerless motion capture,''
\newblock in {\em 2018 International Conference on 3D Vision (3DV)}. IEEE,
  2018, pp. 474--483.

\bibitem{ionescu2013human3}
Catalin Ionescu, Dragos Papava, Vlad Olaru, and Cristian Sminchisescu,
\newblock ``Human3. 6m: Large scale datasets and predictive methods for 3d
  human sensing in natural environments,''
\newblock {\em IEEE transactions on pattern analysis and machine intelligence},
  vol. 36, no. 7, pp. 1325--1339, 2013.

\bibitem{cao2018openpose}
Zhe Cao, Gines Hidalgo, Tomas Simon, Shih-En Wei, and Yaser Sheikh,
\newblock ``Open{P}ose: realtime multi-person 2{D} pose estimation using {P}art
  {A}ffinity {F}ields,''
\newblock in {\em arXiv preprint arXiv:1812.08008}, 2018.

\bibitem{mono-3dhp2017}
Dushyant Mehta, Helge Rhodin, Dan Casas, Pascal Fua, Oleksandr Sotnychenko,
  Weipeng Xu, and Christian Theobalt,
\newblock ``Monocular 3d human pose estimation in the wild using improved cnn
  supervision,''
\newblock in {\em 3D Vision (3DV), 2017 Fifth International Conference on}.
  IEEE, 2017.

\bibitem{Trumble:BMVC:2017}
Matt Trumble, Andrew Gilbert, Charles Malleson, Adrian Hilton, and John
  Collomosse,
\newblock ``Total capture: 3d human pose estimation fusing video and inertial
  sensors,''
\newblock in {\em 2017 British Machine Vision Conference (BMVC)}, 2017.

\bibitem{andriluka14cvpr}
Mykhaylo Andriluka, Leonid Pishchulin, Peter Gehler, and Bernt Schiele,
\newblock ``2d human pose estimation: New benchmark and state of the art
  analysis,''
\newblock in {\em IEEE Conference on Computer Vision and Pattern Recognition
  (CVPR)}, June 2014.

\end{thebibliography}

\end{document}